\documentclass[letterpaper]{article} 
\usepackage[]{aaai2026}  
\usepackage{times}  
\usepackage{helvet}  
\usepackage{courier}  
\usepackage[hyphens]{url}  
\usepackage{graphicx} 
\usepackage{natbib}  
\usepackage{caption} 
\usepackage{algorithm}
\usepackage{algorithmic}

\usepackage{newfloat}
\usepackage{listings}
\DeclareCaptionStyle{ruled}{labelfont=normalfont,labelsep=colon,strut=off} 
\floatstyle{ruled}
\newfloat{listing}{tb}{lst}{}
\floatname{listing}{Listing}
\usepackage[mode=buildnew]{standalone}
\usepackage{subcaption}
\usepackage{tikz}
\usepackage{pgfplots}
\usepackage[dvipsnames]{xcolor}
\usetikzlibrary{fit,matrix,matrix.skeleton,
                arrows.meta, positioning, shapes,
                backgrounds, chains}
\usepackage{amsmath}
\usepackage{booktabs}
\usepackage{multirow}
\usepackage[table]{xcolor}
\nocopyright 

\title{Learning Admissible Heuristics via Cost Partitioning}
\author{
    Hugo Barral\textsuperscript{\rm 1,\rm 2},
    Quentin Cappart\textsuperscript{\rm 2,\rm 3},
    Marie-José Huguet\textsuperscript{\rm 1},
    Sylvie Thiébaux\textsuperscript{\rm 1,\rm 4}
}
\affiliations{
    \textsuperscript{\rm 1}LAAS-CNRS, Université de Toulouse, CNRS, INSA, Toulouse, France \\
    \textsuperscript{\rm 2}Polytechnique Montréal, Montréal, Canada\\
    \textsuperscript{\rm 3}UCLouvain, Louvain-la-Neuve, Belgium\\
    \textsuperscript{\rm 4}Australian National University, Canberra, Australia\\

    \{hugo.barral,marie-jose.huguet\}@laas.fr, quentin.cappart@polymtl.ca, sylvie.thiebaux@anu.edu.au
}

\usepackage{amsmath}
\usepackage{amssymb}
\usepackage{xspace}
\usepackage{todonotes}

\def\hbar#1{\todo[inline,color=brown,size=normalsize]{@hbar: #1}}

\newcommand{\Omit}[1]{}

\newcommand{\texmath}[1]{\ensuremath{#1}\xspace}
\newcommand{\mathsc}[1]{\textrm{\textsc{#1}}}

\newcommand{\task}{\texmath{\Pi}}
\newcommand{\plan}{\texmath{\pi}}
\newcommand{\cost}{\texmath{c}}

\newcommand{\preds}{\texmath{\mathcal{P}}}
\newcommand{\schemas}{\texmath{\mathcal{A}}}
\newcommand{\objects}{\texmath{\mathcal{C}}}
\newcommand{\istate}{\texmath{s_0}}
\newcommand{\goal}{\texmath{G}}
\newcommand{\schparams}[1]{\texmath{\Delta(#1)}}
\newcommand{\pre}[1]{\texmath{{\sf{pre}}(#1)}}
\newcommand{\add}[1]{\texmath{{\sf{add}}(#1)}}
\newcommand{\del}[1]{\texmath{{\sf{del}}(#1)}}
\newcommand{\taskL}{\texmath{\task^\mathsf{lifted}}}
\newcommand{\tupleLifted}{\texmath{\taskL = \langle \preds, \schemas, \objects, \istate, \goal \rangle}}

\newcommand{\vars}{\texmath{V}}
\newcommand{\ops}{\texmath{O}}
\newcommand{\estate}{\texmath{s_*}}
\newcommand{\dom}[1]{\texmath{D_{#1}}}
\newcommand{\eff}[1]{\texmath{{\sf{eff}}(#1)}}
\newcommand{\res}[1]{\texmath{{\sf{succ}}(#1)}}
\newcommand{\taskFDR}{\texmath{\task^\mathsf{FDR}}}
\newcommand{\tupleFDR}{\texmath{\taskFDR = \langle \vars, \ops, \istate, \estate \rangle}}

\newcommand{\sasp}{\text{SAS}\texmath{^+}}

\newcommand{\pattern}{\texmath{P}}
\newcommand{\patternset}{\texmath{\mathcal{S}}}
\newcommand{\opsP}{\texmath{\ops^\pattern}}
\newcommand{\istateP}{\texmath{\istate^\pattern}}
\newcommand{\estateP}{\texmath{\estate^\pattern}}
\newcommand{\taskP}{\texmath{\task^\pattern}}
\newcommand{\tupleP}{\texmath{\taskP = \langle \pattern, \opsP, \istateP, \estateP \rangle}}

\newcommand{\costpartconstraint}[1]{\texmath{\sum_{\pattern \in \patternset} \cost^\pattern (#1) \leq \cost (#1)}}

\newcommand{\graph}{\texmath{G}}
\newcommand{\nodes}{\texmath{N}}
\newcommand{\edges}{\texmath{E}}
\newcommand{\ncolours}{\texmath{\kappa}}
\newcommand{\ecolours}{\texmath{l}}
\newcommand{\setncol}{\texmath{\Sigma_\nodes}}
\newcommand{\setecol}{\texmath{\Sigma_\edges}}
\newcommand{\neighboors}{\texmath{\mathcal{N}}}

\newcommand{\FLG}{\texmath{\mathsc{FLG}}}
\newcommand{\graphFLG}{\texmath{\graph_\FLG}}
\newcommand{\evarval}{\texmath{\edges_{\texttt{var:val}}}}
\newcommand{\epre}{\texmath{\edges_{\texttt{pre}}}}
\newcommand{\eeff}{\texmath{\edges_{\texttt{eff}}}}

\newcommand{\AOAG}{\texmath{\mathsc{AOAG}}}
\newcommand{\graphAOAG}{\texmath{\graph_\AOAG}}

\newcommand{\iters}{\texmath{L}}

\newcommand{\X}{\texmath{\mathbf{X}}}
\newcommand{\odim}{\texmath{m}}
\newcommand{\pdim}{\texmath{k}}
\newcommand{\vdim}{\texmath{d}}
\newcommand{\Xdims}[1]{\texmath{\X(#1) \in \mathbb{R}^{\pdim \times \odim \times \vdim}}}
\newcommand{\carray}{\texmath{\mathbf{H}}}
\newcommand{\weights}{\texmath{\mathbf{\alpha}}}
\newcommand{\loss}{\texmath{\mathcal{L}}}

\newcommand{\hLCP}{\texmath{h^{\mathsc{LCP}}}}
\newcommand{\hLCPRL}[2]{\texmath{h^{\mathsc{LCP}}_\mathsc{#1-#2}}}
\newcommand{\hOCP}{\texmath{h^{\mathsc{OCP}_+}}}
\newcommand{\hSCP}{\texmath{h^{\mathsc{SCP}_+}}}
\newcommand{\hGZOCP}{\texmath{h^{\mathsc{GZOCP}}}}

\definecolor{thirdgray}{gray}{0.85} 
\definecolor{secondgray}{gray}{0.7}  
\definecolor{bestgray}{gray}{0.6}  
\definecolor{tri_blue}{RGB}{80,80,160}
\definecolor{tri_green}{RGB}{80,160,80}
\definecolor{tri_ora}{RGB}{245,73,39}
\definecolor{mpiblue}{HTML}{33a5c3}

\begin{document}

\maketitle

\begin{abstract}
Admissible heuristics are essential for optimal planning, yet learning them remains challenging due to the risk of overestimation. Cost partitioning combines multiple abstraction heuristics while preserving admissibility, but computing optimal partitions online is expensive. We propose a framework that learns to infer admissible cost partitions by leveraging the Lagrangian dual equivalence between cost partitioning and multiplier prediction. Planning states and patterns are encoded as labelled graphs, and an action‑centric variant of the Weisfeiler–Leman algorithm extracts structural feature vectors. A deep architecture with axial self‑attention and a softmax output layer maps these features to cost weights that satisfy the partition constraints by construction, ensuring admissibility. Experiments demonstrate reduced node expansions compared to suboptimal partitioning baselines while maintaining strict admissibility. 
To our knowledge, this is the first machine-learned heuristic guaranteed to be admissible.
\end{abstract}

\section{Introduction}
Recent advances in machine learning are starting to have an impact on
planning, in particular through learning search guidance in the form of heuristic functions \cite{shen:etal:20,chen:etal:24:icaps,horcik:etal:25,correa:etal:25}, state rankings \cite{garrett:etal:16,hao:etal:25}, policies \cite{toyer:etal:20,stahlberg:etal:22:icaps,stahlberg:etal:22:kr,silver:etal:24,chen:etal:25}, and action pruning \cite{musayev:etal:25}. 
However, the focus of these works has been on improving the scalability of planning, rather than the quality of the solutions. In particular, learning admissible heuristics, able to guide the search of optimal planners, remains an open problem. Some contributions present models that achieve low residual errors or relaxed probabilistic guarantees \cite{ernandes:gori:04,marom:rosman:20,futuhi:stuertevant:26}.
Yet, despite their accuracy, these learned estimators lack formal admissibility guarantees. Since a single overestimation of $h^*$ can cause A* to return a suboptimal plan, the landscape still lacks a robust method to learn admissible heuristics for optimal planning.

To address this challenge, we explore a new research direction. Instead of directly attempting to learn  admissible heuristics from scratch, we learn to combine admissible heuristics via cost partitioning. Cost partitioning is a well-established, rigorous method which
distributes the cost of actions amongst multiple admissible heuristics, to ensure that adding them
remains admissible, thereby producing more informed admissible heuristics \cite{haslum:etal:05,katz:domshlak:10,pommerening:etal:13,pommerening:etal:15,seipp:etal:20}.
For certain classes of admissible heuristics such as abstraction heuristics, an optimal cost partitioning (OCP)
maximizing the combined heuristic value over all partitions can be computed in polynomial time via linear programming \cite{katz:domshlak:10}.
However the resulting linear program (LP) is too large to be solved at every node of the search, which reduces the
usage of OCP in practice. This suggests that a learning-based approach could be beneficial to speed up the process.

  
  
  


In this paper, we develop such an approach, taking inspiration from two key results from the literature.
The first is the theoretical insight that OCP is equivalent to a Lagrangian dual problem~\cite{pommerening:etal:19}:
the multipliers that maximize a Lagrangian lower bound, and provide tight bounds in branch‑and‑bound, directly correspond to cost partition weights.
In other words, each multiplier encodes how much of an action’s original cost is allocated to a particular abstraction.
The second result is the recent development of machine learning models that can be trained to predict tight Lagrangian multipliers for integer linear and constraint programs \cite{abbas:swoboda:24,parjadis:etal:24,bessa:etal:25}.
Taken together, these results suggest it should be possible to develop similar machine learning approaches to infer valid cost partitions.


We introduce a novel framework that is based on this perspective. Rather than solving an optimization problem from scratch for every evaluated state, we train a domain-specific model capable of inferring valid cost partitions online.
Our framework constructs a graph representation of a planning state 
and its abstractions, from which we extract Weisfeiler-Leman (WL) feature vectors \cite{chen:etal:24:icaps}.
A deep learning architecture then contextualizes the feature vectors via a self-attention mechanism~\cite{vaswani:etal:17} and produces cost partition weights. 
A softmax output layer enforces that the weights sum to 1,
ensuring the resulting heuristic remains admissible.
The model is trained on optimal cost partitions from various planning instances from the given domain, in order to estimate a function that compute optimal cost partitions from contextualized action features.
Once learned, this relationship allows the model to infer effective cost partitions even on problems larger than those encountered during training.

We empirically demonstrate through experiments on 6 domains that these inferred partitions yield a strictly admissible heuristic that can reduce search node expansions compared to suboptimal cost partitioning baselines.

The remainder of this paper is structured as follows. Section 2 provides the necessary background and notation for our learning approach to planning. Section 3 details our model architecture for inferring cost partition. Section 4 presents an empirical evaluation on 6 International Planning Competition domains, with a focus on coverage, node expansions, and search time. Section 5 concludes with a discussion of limitations and directions for future work.

\section{Background and Notation}
\subsection{Planning}\label{sec:planning}
Throughout this paper we assume that planning tasks are provided in lifted form and internally grounded and transformed into a finite domain representation (FDR) by the planner prior to search \cite{helmert:09}. We define these two representations below.

A lifted planning task is a tuple \tupleLifted, where \preds is a set of predicates, \schemas is a set of action schemas, \objects is a set of
{object constants}, \istate is the initial state and \goal is the goal condition.
A predicate $p \in \preds$ {of arity $n_p\in \mathbb{N}_0$ has parameters $\{x_i \ | \ 1 \leq i \leq n_p \}$.}
An action schema $a \in \schemas$ {of arity $n_a\in \mathbb{N}_0$} is defined by a tuple $\langle \schparams{a}, \pre{a}, \add{a}, \del{a} \rangle$, where \schparams{a} is a set of parameters {such that $|\schparams{a}|=n_a$}, and \pre{a}, \add{a} and \del{a} are sets of predicates from \preds instantiated with either parameter variables or objects in $\schparams{a} \cup \objects$.
A (ground) atom
(resp. action) is an instance of a predictate (resp. action schema) obtained by substituting all its parameters with objects from \objects. 
A state in the lifted representation is
{the set of all} atoms
that are true in that state.
The goal $G$ is a set of 
atoms and a state $s$ is a goal state if $\goal \subseteq s$.

\Omit{
A lifted planning task is a tuple $\task = \langle \preds, \schemas, \objects, \istate, \goal \rangle$, where \preds is a set of predicates, \schemas is a set of action schemas, \objects is a set of constant objects, \istate is the initial state and \goal is the goal condition.
A predicate $P \in \preds$ has $\{x_i \ | \ 1 \leq i \leq n_P \}$ parameters for $n_P \in \mathbb{N}_0$.
A predicate can be instantiated as a proposition by assigning its parameters with objects from \objects.
An action schema $a \in \schemas$ is defined by a tuple $\langle \schparams{a}, \pre{a}, \add{a}, \del{a} \rangle$, where \schparams{a} is a set of parameters and \pre{a}, \add{a} and \del{a} are sets of predicated from \preds instantiated with either parameter variables or objects in $\schparams{a} \cup \objects$.
An action schema can be instantiated as an action by assigning all its parameters with objects from \objects.
A predicate (resp. an action schema) with $n$ parameters is an $n$-ary predicate (resp. action schema).
A state in the lifted representation is a set of propositions that are true in that state.
A state $s$ is a goal state if $G \subseteq s$.
}

An FDR planning task is a tuple \tupleFDR, where $V$ is a set of finite-domain state variables, $O$ is a set of actions, \istate is the initial state and \estate is the goal {condition}.
Each $v \in \vars$ has a finite domain \dom{v}. 
A fact is a pair $\langle v, d \rangle$, with $v \in \vars, d \in \dom{v}$, and equates a ground 
atom of the underlying lifted task.
An assignment (resp. a partial assignment) is a set of facts where {each} $v \in V$ appears once (resp. at most once).
Each $o \in O$ is defined by its precondition \pre{o} and effect \eff{o}, expressed as partial assignments on $V$.
Finally, 
$s_0$ and $s_*$ are {an} assignment and partial assignment, respectively.

\Omit{A grounded planning task $\task = \langle \vars, \ops, \istate, \estate \rangle$, where $V$ is a set of finite-domain state variables, $O$ is a set of actions, \istate is the initial state and \estate is the goal state.
Each $v \in \vars$ has a finite domain \dom{v}, derived from grounded predicates.
A fact is a pair $\langle v, d \rangle, v \in \vars, d \in \dom{v}$.
An assignment (resp. a partial assignment) is a set of facts where $\forall v \in V$ appears once (resp. at most once).
Each $o \in O$ is defined with precondition \pre{o} and effect \eff{o} expressed as partial assignments on $V$.
$s_0$ and $s_*$ are respectively assignment and partial assignment.}

An action $o$ is applicable in state $s$ if $\pre{o} \subseteq s$, and leads to the successor state $\res{s,o} = s \oplus \eff{o}$ where $\oplus$ replaces $(v,d)\in s$ with $(v,d')\in \eff{o}$ if $d \neq d'$.
A solution or a plan is a sequence of actions $\plan = \langle o_1, \dots, o_n \rangle$, that can be applied to $s_0$ and to all the subsequent states to achieve a goal state.
That is, $\pi$ induces a sequence of states $s_0 , \dots , s_n$ such that $\pre{o_i} \subseteq s_{i-1}$ and $s_{i} = \res{s_{i-1},o_i}$ for $i= 1,\dots, n$, and $\estate \subseteq s_n$.
An action cost is defined by $\cost (o) \in \mathbb{R}_0^+$.
The cost of \plan is $\cost(\plan) = \sum_{o_i \in \plan} \cost(o_i)$.
A planning task is solvable if there exists at least one plan.
In an optimal planning setting, we would like to find a plan $\pi^*$ of minimum cost, i.e. $\arg\min_{\pi^*} \cost (\pi^*)$.

\subsection{Abstraction Heuristics and Cost Partitioning}\label{sec:abs_cost_part}
Abstraction heuristics map states to a smaller set of abstract states, treating concrete states mapped to the same abstract state as equivalent under the abstraction.
This
induces an abstract transition system whose size is much smaller than the original state space.
In this abstract system, optimal goal distances yield an admissible heuristic for the original task and can be computed efficiently using uninformed search algorithms.
Abstractions are commonly generated by projecting tasks onto a small set of variables or \textit{pattern} $\pattern \subseteq \vars$.
We write \tupleP for the pattern projection of \taskFDR, where
for a (partial) assignment $s$, the projection $s|_P$ is the restriction of $s$ to the variables in $P$, and 
for each $o \in O$, the projected action $o^\pattern \in \opsP$ keeps $\pre{o^\pattern} = \pre{o}|_\pattern$ and $\eff{o^\pattern} = \eff{o}|_\pattern$.

\Omit{
Abstraction heuristics reduce the state-space by mapping concrete states to a smaller set of abstract states, distinct concrete states are treated as equivalent under the abstraction. This method induces an abstract transition system whose size is much smaller than the original state-space.
In this abstract system, optimal goal distances can be computed efficiently using algorithms such as Dijkstra~\cite{Dijkstra}, yielding an admissible heuristic for the original task.
Pommerening et al.~\cite{} provide a general representation of abstraction as pattern. 
Let $\pattern \subseteq \vars$ be a pattern that define a subset of a task variables.
We note $\Pi^P = \langle P, O^P, s^P_0, s_*^P \rangle$, a pattern projection of \task.
States are restricted to variables in \pattern.
For each $o \in O$, the projected action $o^P$ keeps $\pre{o^P} = \pre{o}|_P$ and $\eff{o^P} = \eff{o}|_P$.
}

To focus on multiple aspects of the task and obtain a more informed heuristic estimate, multiple abstractions $\pattern_1, \dots, \pattern_\pdim$ are typically combined.
However, these heuristics are generally not additive,
as the same action may be applied redundantly across multiple abstractions.
Cost partitioning addresses this by distributing the cost of each action among the abstractions before computing their heuristics, ensuring that the sum of the per-abstraction heuristic estimates never exceeds the true cost~\cite{katz:domshlak:10}.
Let us note $\cost^P$, the cost function distributed to a pattern abstraction $P$.
For the distribution to ensure the additivity of a set of patterns \patternset, the following constraint must be satisfied:
\begin{equation}
  \label{eqn:cost-partitioning-constraint}
  \costpartconstraint{o}, \forall o \in O
\end{equation}
If each abstraction $P$ is evaluated under its assigned cost function $\cost^P$, producing heuristic $h^P$, then the additive heuristic
\begin{equation}
  \label{eqn:additive-heuristic}
  h^\patternset(s, \cost) = \sum_{P \in \patternset} h^P(s, \cost^P)
\end{equation}
is admissible at state $s$.

The informativeness
of the resulting additive heuristic depends on the chosen abstractions and partition, and also on the current search state $s$.
Optimal Cost Partitioning (OCP) is the the problem of computing the cost partitioning that maximises the heuristic value $h^\patternset(s,\cost)$. It can be can be expressed as an LP that additively combines abstraction heuristics as in~(\ref{eqn:additive-heuristic}) by enforcing the cost partition constraint~(\ref{eqn:cost-partitioning-constraint})~\cite{pommerening:etal:14}. However, solving this LP anew at each state is
%
computationally prohibitive in practice.
Existing approaches either settle for faster, suboptimal partitioning
~\cite{karpas:domshlak:09,seipp:etal:17} and/or do not recompute the partition at every state~\cite{seipp:21}.

Furthermore, a theoretical result by \citet{pommerening:etal:19} shows that the OCP problem, whose dual belongs to the operator-counting framework~\cite{pommerening:etal:14}, can be seen as maximizing a sum of independent sub‑problems, each parameterized by Lagrangian multipliers.
Consequently, any technique capable of producing Lagrangian multipliers also yields a valid cost partition, providing an alternative avenue to tackle the online computational challenge.
In particular, \citet{parjadis:etal:24} showed that machine learning can be used to obtain tight multipliers in the context of Lagrangian relaxation applied to constraint programming, while \citet{abbas:swoboda:24,bessa:etal:25} demonstrated similar results for Lagrangian decomposition. Inspired by this work, we propose a related approach to learn Lagrangian multipliers that yield valid and tight cost partitions.

\subsection{Graph Representation of Planning States}\label{sec:graph_rep}
\begin{algorithm}[tb]
\caption{WL algorithm}
\label{alg:wl}
\textbf{Input}: Graph $\graph = \langle \nodes, \edges, \ncolours, \ecolours \rangle$; number of iterations \iters.\\
\textbf{Output}: Set of colours.
\begin{algorithmic}[1] 
\STATE $\ncolours^{0}(v) \leftarrow \ncolours(v), \forall v \in V$ \label{line:wl:init}\\ 
\FOR{$j=1,\ldots,\iters$ \normalfont{\textbf{do for}} $v \in V$}
\STATE $\ncolours^{j}(v) \leftarrow \text{hash} \big(\ncolours^{j-1}(v), \{(\ncolours^{j-1}(u), \iota) \mid \iota \in \Sigma_E, u \in \neighboors_\iota (v)\} \big)$
\label{line:wl:update}
\ENDFOR
\STATE \textbf{return} {$\bigcup_{j=0,\ldots,L}\{\ncolours^{j}(v) \mid v \in V\}$} \label{line:wl:return}
\end{algorithmic}
\end{algorithm}

A suitable learning approach must start with an encoding of planning states.
A graph representation is a natural choice because a planning state fundamentally consists of entities connected by structural relationships.
A labelled graph is a tuple $\graph = \langle \nodes, \edges, \ncolours, \ecolours \rangle$, where \nodes is a set of nodes, $\edges \subseteq \binom{\nodes}{2}$ is a set of undirected edges, $\ncolours : \nodes \rightarrow \setncol$ maps nodes to a set of colours \setncol, and $\ecolours : \edges \rightarrow \setecol$ maps edges to a set of colours \setecol.
The neighbourhood of a node $u$ under edge colour $\iota$ is $\neighboors_\iota (u) = \{v \mid (u,v) \in \edges \cap \ecolours(u,v) = \iota \}$.
The neighbourhood of a node $u$ in a graph is $\neighboors (u) = \bigcup_{\iota \in \setecol} \neighboors_\iota (u)$.

We rely on the modified Weisfeiler–Leman (WL) algorithm~\cite{chen:etal:24:icaps} to extract structural features.
The original WL algorithm is an incomplete test to decide graph non-isomorphism.
It has since  evolved into a reliable method for extracting a vector representation of a graph's structure~\cite{shervashidze:etal:11}.
Because the WL procedure is entirely deterministic, it provide a consistent and reproducible representation of each state‑pattern pair, suitable as input to a learned predictor.

Alg.~\ref{alg:wl} shows the WL algorithm that takes as input a labelled graph \graph and a number of iteration \iters.
After initializing the nodes original colour, Line~\ref{line:wl:update} iteratively updates the colours of each node $v$.
It aggregates the unions of node and edge colours of its neighbours in a set and hash them with $v$'s current colour into a colour for the next iteration.
The algorithm returns a set of colours seen over all iterations.
Those set of colours can be represented as histogram vector $v$ with a size \vdim equal to the number of observed colours on a training set of graphs.
Let $v[c]$ count how many times the WL algorithm has encountered colour $c$ throughout its iterations.
As there is no guarantee that all possible colours can be observed from a limited training set of a given planning domain, non-observed colours encountered after training are purely ignored.

\begin{figure*}[tb]
    \centering
    \begin{tikzpicture}[
        element/.style={rectangle, draw=black, fill=cyan!20, very thick, minimum size=5mm},
        process/.style={rectangle, draw=red, fill=red!20, very thick, minimum size=5mm},
        arrow/.style={->, line width=0.3mm}
        ]
        
        \node[element] (state) {state $s$};
        \node[process, below=1em of state] (generate) {Generate graphs};
        \node[element, below=1em of generate] (patterns) {patterns \patternset};
        
        \node[right=1em of generate] (output_graph) {
        \scalebox{0.55} {
            \begin{tikzpicture}
            \node (original_graph) {
                \begin{tikzpicture}[
                    node distance=2em and 1em,
                    every node/.style={circle, draw, minimum size=1.2em, inner sep=2pt, font=\small},
                    bnode/.style={fill=tri_blue},
                    gnode/.style={fill=tri_green},
                    rnode/.style={fill=tri_ora},
                    dots/.style={draw=none, circle=False, minimum size=0pt}
                ]
            
                \node[bnode] (L1) {};
                \node[bnode, right=of L1] (L2) {};
                \node (Ldots) [right=of L2, dots] {$\dots$};
                \node[bnode] (LN) [right=of Ldots] {};
            
                \node[gnode] (M1) [below=of L1] {};
                \node[gnode] (M2) [right=of M1] {};
                \node (Mdots) [right=of M2, dots] {$\dots$};
                \node[gnode] (MN) [right=of Mdots] {};
            
                \node[rnode] (O1) [below=of M1] {$o_1$};
                \node[rnode] (O2) [below=of M2] {$o_2$};
                \node (Odots) [below=of Mdots, dots] {$\dots$};
                \node[rnode] (ON) [below=of MN] {$o_\odim$};
            
                \draw (L1) -- (MN);
                \draw (L2) -- (M1);
                \draw (L2) -- (M2);
                \draw (LN) -- (M1);
                \draw (LN) -- (MN);
            
                \draw (M1) -- (O1);
                \draw (M1) -- (O2);
                \draw (M2) -- (O1);
                \draw (M2) -- (ON);
                \draw (MN) -- (O1);
                \draw (MN) -- (ON);
            
                \end{tikzpicture}
            };

            \node[right=0.2em of original_graph, yshift=7.8em, rotate=270] (set_graph) {$\underbrace{\hspace{15em}}$};
            
            \node[right=1em of set_graph, yshift=15em] (pattern_graph1) {
                \scalebox{0.5} {
                \begin{tikzpicture}[
                    node distance=2em and 1em,
                    every node/.style={circle, draw, minimum size=1.2em, inner sep=2pt, font=\small},
                    bnode/.style={fill=tri_blue},
                    gnode/.style={fill=tri_green},
                    rnode/.style={fill=tri_ora},
                    mask/.style={dashed, opacity=0.2},
                    dots/.style={draw=none, circle=False, minimum size=0pt}
                ]
            
                \node[bnode, mask] (L1) {};
                \node[bnode, right=of L1] (L2) {};
                \node (Ldots) [right=of L2, dots] {$\dots$};
                \node[bnode, mask] (LN) [right=of Ldots] {};
            
                \node[gnode] (M1) [below=of L1] {};
                \node[gnode] (M2) [right=of M1] {};
                \node (Mdots) [right=of M2, dots] {$\dots$};
                \node[gnode, mask] (MN) [right=of Mdots] {};
            
                \node[rnode] (O1) [below=of M1] {$o_1$};
                \node[rnode] (O2) [below=of M2] {$o_2$};
                \node (Odots) [below=of Mdots, dots] {$\dots$};
                \node[rnode] (ON) [below=of MN] {$o_\odim$};
            
                \draw[mask] (L1) -- (MN);
                \draw (L2) -- (M1);
                \draw (L2) -- (M2);
                \draw[mask] (LN) -- (M1);
                \draw[mask] (LN) -- (MN);
            
                \draw (M1) -- (O1);
                \draw (M1) -- (O2);
                \draw (M2) -- (O1);
                \draw (M2) -- (ON);
                \draw[mask] (MN) -- (O1);
                \draw[mask] (MN) -- (ON);
            
                \end{tikzpicture}
                }
            };
            \node[right=0.25em of pattern_graph1, scale=1.5] (pattern2) {$\pattern_1$};
            
            \node[below=0.25em of pattern_graph1] (pattern_graph2) {
                \scalebox{0.5} {
                \begin{tikzpicture}[
                    node distance=2em and 1em,
                    every node/.style={circle, draw, minimum size=1.2em, inner sep=2pt, font=\small},
                    bnode/.style={fill=tri_blue},
                    gnode/.style={fill=tri_green},
                    rnode/.style={fill=tri_ora},
                    mask/.style={dashed, opacity=0.2},
                    dots/.style={draw=none, circle=False, minimum size=0pt}
                ]
            
                \node[bnode] (L1) {};
                \node[bnode, mask, right=of L1] (L2) {};
                \node (Ldots) [right=of L2, dots] {$\dots$};
                \node[bnode, mask] (LN) [right=of Ldots] {};
            
                \node[gnode, mask] (M1) [below=of L1] {};
                \node[gnode, mask] (M2) [right=of M1] {};
                \node (Mdots) [right=of M2, dots] {$\dots$};
                \node[gnode] (MN) [right=of Mdots] {};
            
                \node[rnode] (O1) [below=of M1] {$o_1$};
                \node[rnode] (O2) [below=of M2] {$o_2$};
                \node (Odots) [below=of Mdots, dots] {$\dots$};
                \node[rnode] (ON) [below=of MN] {$o_\odim$};
            
                \draw (L1) -- (MN);
                \draw[mask] (L2) -- (M1);
                \draw[mask] (L2) -- (M2);
                \draw[mask] (LN) -- (M1);
                \draw[mask] (LN) -- (MN);
            
                \draw[mask] (M1) -- (O1);
                \draw[mask] (M1) -- (O2);
                \draw[mask] (M2) -- (O1);
                \draw[mask] (M2) -- (ON);
                \draw (MN) -- (O1);
                \draw (MN) -- (ON);
            
                \end{tikzpicture}
                }
            };
            \node[right=0.25em of pattern_graph2, scale=1.5] (pattern2) {$\pattern_2$};
            
            \node[below=0.25em of pattern_graph2, scale=1.5] (vdots) {$\vdots$};
            
            \node[below=0.25em of vdots] (pattern_graphM) {
                \scalebox{0.5} {
                \begin{tikzpicture}[
                    node distance=2em and 1em,
                    every node/.style={circle, draw, minimum size=1.2em, inner sep=2pt, font=\small},
                    bnode/.style={fill=tri_blue},
                    gnode/.style={fill=tri_green},
                    rnode/.style={fill=tri_ora},
                    mask/.style={dashed, opacity=0.2},
                    dots/.style={draw=none, circle=False, minimum size=0pt}
                ]
            
                \node[bnode, mask] (L1) {};
                \node[bnode, mask, right=of L1] (L2) {};
                \node (Ldots) [right=of L2, dots] {$\dots$};
                \node[bnode] (LN) [right=of Ldots] {};
            
                \node[gnode] (M1) [below=of L1] {};
                \node[gnode, mask] (M2) [right=of M1] {};
                \node (Mdots) [right=of M2, dots] {$\dots$};
                \node[gnode] (MN) [right=of Mdots] {};
            
                \node[rnode] (O1) [below=of M1] {$o_1$};
                \node[rnode] (O2) [below=of M2] {$o_2$};
                \node (Odots) [below=of Mdots, dots] {$\dots$};
                \node[rnode] (ON) [below=of MN] {$o_\odim$};
            
                \draw[mask] (L1) -- (MN);
                \draw[mask] (L2) -- (M1);
                \draw[mask] (L2) -- (M2);
                \draw (LN) -- (M1);
                \draw (LN) -- (MN);
            
                \draw (M1) -- (O1);
                \draw (M1) -- (O2);
                \draw[mask] (M2) -- (O1);
                \draw[mask] (M2) -- (ON);
                \draw (MN) -- (O1);
                \draw (MN) -- (ON);
            
                \end{tikzpicture}
                }
            };
            \node[right=0.25em of pattern_graphM, scale=1.5] (patternM) {$\pattern_\pdim$};
            
        \end{tikzpicture}
        }
        };
        
        \node[process, right=1em of output_graph] (extract) {Feature extract};
        
        \node[right=1em of extract, xshift=-0.4em] (features) {
        \begin{tikzpicture}
            \coordinate (O) at (0,0,0);
            \coordinate (A) at (0,2,0);
            \coordinate (B) at (0,2,2);
            \coordinate (C) at (0,0,2);
            \coordinate (D) at (2,0,0);
            \coordinate (E) at (2,2,0);
            \coordinate (F) at (2,2,2);
            \coordinate (G) at (2,0,2);
            \draw[mpiblue!60!black,fill=mpiblue!5] (O) -- (C) -- (G) -- (D) -- cycle;
            \draw[mpiblue!60!black,fill=mpiblue!5] (O) -- (A) -- (E) -- (D) -- cycle;
            \draw[mpiblue!60!black,fill=mpiblue!5] (O) -- (A) -- (B) -- (C) -- cycle;
            \draw[mpiblue!60!black,fill=mpiblue!5,opacity=0.8] (D) -- (E) -- (F) -- (G) -- cycle;
            \draw[mpiblue!60!black,fill=mpiblue!5,opacity=0.6] (C) -- (B) -- (F) -- (G) -- cycle;
            \draw[mpiblue!60!black,fill=mpiblue!5,opacity=0.8] (A) -- (B) -- (F) -- (E) -- cycle;
             \coordinate (O) at (1,0.85,1.5);  
             
            \draw (0,-1.3,0) node {\scriptsize{\odim}}; 
            \draw (-0.9,-1,0) node {$\underbrace{\hspace{2cm}}$}; 
            \draw (-0.6,1,2) node {\scriptsize{\pdim}}; 
            \draw (-0.1,2.1,2) node[rotate = 270] {$\underbrace{\hspace{2cm}}$}; 
            \draw (2.5,0,1.2) node[below right = -0.1cm and -0.3cm] {\scriptsize{\vdim}}; 
            \draw (2.2,0,2.5) node[rotate = 45] {$\underbrace{\hspace{1.1cm}}$}; 
            \draw (O) node {\X}; 
            
        \end{tikzpicture}
        };
        
        \node[process, right=1em of features] (model) {Model};
        \node[right=1em of model, yshift=-0.75em] (partition) {
        \begin{tikzpicture}
           
            \coordinate (O) at (0.8,0.2);
            \coordinate (A) at (0.8,1.2);
            \coordinate (B) at (1.8,0.2);
            \coordinate (C) at (1.8,1.2);
            \draw[fill=black!5] (O) -- (A) -- (C) -- (B) --cycle;
             \coordinate (Y) at (1.1,0.75);  
             
            \draw (1.9,0.7) node {\scriptsize{\pdim}}; 
            \draw (1.9,0.1) node[rotate = 90] {$\underbrace{\hspace{1cm}}$}; 
            \draw (1,-0.2) node {\scriptsize{\odim}}; 
            \draw (0.7,0.1) node {$\underbrace{\hspace{1cm}}$}; 
            \draw (Y) node {\weights}; 
            
        \end{tikzpicture}
        };
        
        \draw[arrow] (state.south) -- (generate.north);
        \draw[arrow] (patterns.north) -- (generate.south);
        \draw[arrow] (generate.east) -- (output_graph.west);
        \draw[arrow] (output_graph.east) -- (extract.west);
        \draw[arrow] (extract.east) -- ([xshift=0.5em]features.west);
        \draw[arrow] ([xshift=-1em]features.east) -- (model.west);
        \draw[arrow] (model.east) -- ([yshift=0.75em, xshift=0.75em]partition.west);

    \end{tikzpicture}
    \caption{Three‑stage pipeline for learning cost partitions. (Left) a state and its patterns are encoded as labelled graphs. (Center) WL algorithm yields action‑centric embeddings arranged in a 3D array. (Right) Model with self‑attention mechanism predicts admissible partition weights. \pdim denotes the number of patterns, \odim the number of actions and \vdim the number of features.}
    \label{fig:overview}
\end{figure*}
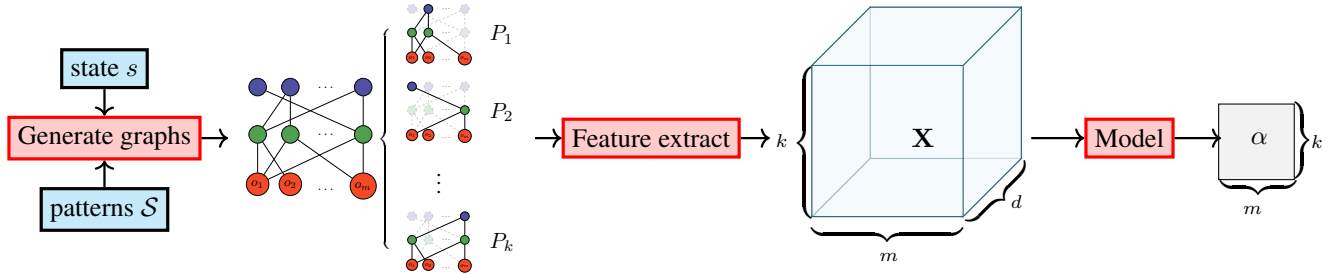

\section{Learning to Infer Cost Partitions}

In this section, we describe our methodology to avoid solving the optimal cost partitioning problem from scratch for every evaluated state.
We define a framework to learn a domain‑specific model that predicts a valid cost partition for the current state and a set of abstractions.
Figure~\ref{fig:overview} schematizes the full pipeline.
First, the planner state and each pattern are jointly encoded as a labelled graph.
From there, we leverage a variant of the WL algorithm to extract action‑centric feature vectors; these are shaped into a three‑dimensional array whose axes correspond to abstraction, ground action, and feature dimension.
Finally, a neural network equipped with self‑attention contextualizes the feature array and, via a softmax output layer, produces cost partition weights that satisfy the sum constraint.
Section~\ref{sec:feature_extract} details the graph representations and our method to project graph to abstraction patterns.
Section~\ref{sec:action_wl} presents our WL variant to extract action-centric features.
Section~\ref{sec:model_architecture} describes the attention‑based predictor and how this model enforces admissibility.

\subsection{Graph Representation of Abstractions}
\label{sec:feature_extract}

As described in Section~\ref{sec:graph_rep}, a planning state $s$ must be encoded as a labelled graph $\graph(s)$ that captures the entities and structural relationships of the planning task.
We adapt two graph representations $\graph (s)$ from the literature, the Action-Object-Atom graph (AOAG) \cite{wang:trevizan:25} and the FDR learning graph (FLG) \cite{chen:etal:24:aaai}, to capture the current task state and the connections between actions and state components.
The former focuses on the lifted structure of objects and atoms, whereas the latter encapsulates a grounded (FDR) view of the task.
We assume our original lifted task is grounded through a translation process that allows us to keep information from both representations.
Let $\mathcal{D}_\task = \bigcup_{v \in \vars} \dom{v}$ be the set of all domain values of a task \task.
During grounding we retain the mappings $\rho: \mathcal{D}_\Pi \rightarrow \preds$ and $sch: \ops \rightarrow \schemas$ that link grounded values and actions to their lifted counterparts.
We write an example planning task $\task_X$ with lifted and grounded elements in Table~\ref{tab:example_task} to illustrate the graph representations.

\begin{table}
    \centering
    \begin{tabular}{c|c}
        Lifted & FDR \\
        \midrule
        $\preds = \{p_1, p_2\}$ & $\vars = \{v_1,v_2\}$\\
        $\objects = \{c_1,c_2\}$ & $D_{v_1} = \{F_1, T_1\}$ \\
        & $D_{v_2} = \{ F_2, T_2\}$\\
        \midrule
        $\schemas = \{a_1(c_1, c_2), a_2(c_1)\}$ & $\ops = \{o_1, o_2\}$ \\
        $\pre{a_1} = \{\neg p_1(c_1), p_2(c_2)\}$& $\pre{o_1} = \{\langle v_1, F_1 \rangle,$\\
        $\pre{a_2} = \{\neg p_2(c_2)\}$& $\langle v_2, T_2 \rangle\}$ \\
        $\add{a_1} = \{p_1(c_1)\}$& $\pre{o_2} = \{\langle v_2, F_2 \rangle\}$\\
        $\add{a_2} = \{p_2(c_2)\}$& $\eff{o_1} = \{\langle v_1, T_1 \rangle,$\\
        $\del{a_1} = \{\neg p_2(c_2)\}$& $\langle v_2, F_2 \rangle\}$\\
        $\del{a_2} = \{\}$& $\eff{o_2} = \{\langle v_2, T_2 \rangle\}$\\
        \midrule
        $\goal = \{p_1(c_1), p_2(c_2)\}$ & $\estate = \{\langle v_1, T_1 \rangle,$\\
        & $\langle v_2, T_2 \rangle\}$\\
        \midrule
        $s = \{p_1(c_1)\}$ & $s = \{\langle v_1, T_1 \rangle, $\\
        & $\langle v_2, F_2 \rangle\}$\\
    \end{tabular}
    \caption{Example task $\task_X$ in lifted and FDR formalisms. The facts $\langle v_1,T_1 \rangle$, $\langle v_2,T_2 \rangle$, $\langle v_1,F_1 \rangle$, $\langle v_2,F_2 \rangle$ in the FDR representation encode the atoms $p_1(c_1)$, $p_2(c_2)$ and their respective negation in the lifted representation.}
    \label{tab:example_task}
\end{table}

\subsubsection{Action-Object-Atom graph}\label{sec:aoag}
We use a particular case of AOAG from~\citet{wang:trevizan:25} where the graph always represent all grounded actions in \ops.
The graph's nodes represent the lifted task's objects, actions, and atoms that are true in the current state or in the goal.
The action nodes are labeled by their corresponding action schema, and the atom nodes are labeled by their ``goal status'' which encodes whether the atom is an achieved goal (\texttt{ag}), an unachieved goal (\texttt{ug}) or a non-goal atom (\texttt{ap} for ``achieved proposition'' since it must then be true in the current state).
The graph edges connect object nodes to the nodes representing the proposition and actions that include this object in their argument list at some position $i$, and are labeled by the index $i$.
Formally, the AOAG of a task \task is the graph $\graphAOAG (s) = \langle \nodes, \edges, \ncolours, \ecolours \rangle$ with:
\begin{itemize}
    \item $\nodes = \objects \cup s \cup \goal \cup \ops$
    \item $\edges = \bigcup_{p(x_1, \dots, x_{n_p}) \in \istate \cup \goal} \{\langle p, x_1 \rangle, \dots, \langle p, x_{n_p} \rangle\} \cup \bigcup_{\schparams{a} = (x_1, \dots, x_{n_a}) \in \schemas} \{\langle a, x_1 \rangle, \dots, \langle a, x_{n_a} \rangle\}$
    \item $\ncolours: V \rightarrow \{\texttt{ob}\} \cup (\{\texttt{ap},\texttt{ag},\texttt{ug}\} \times \preds) \cup \schemas$ defined by 
        \begin{equation*}
            u \rightarrow \begin{cases}
            \texttt{ob}, & \text{if $u \in \objects$}.\\
            (\texttt{ap}, p), & \text{if $u = p \in s \setminus \goal$}.\\
            (\texttt{ag}, p), & \text{if $u = p \in s \ \cap \goal$}.\\
            (\texttt{ug}, p), & \text{if $u = p \in \goal \setminus s$}.\\
            sch(u), & \text{if $u \in O$}.\\
            \end{cases}
        \end{equation*}
      \item $\ecolours: E \rightarrow \mathbb{N}$ with $\ecolours(e) = i$ for $e=\langle \_, x_i \rangle$
\end{itemize}

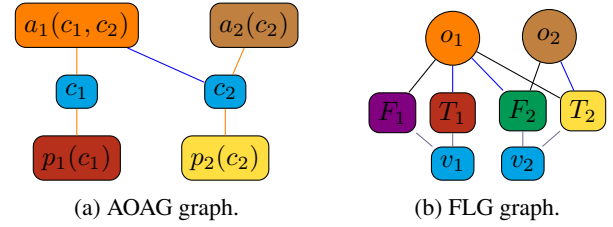
\begin{figure}[tb]
    \centering
        \begin{subfigure}[b]{0.48\linewidth}
            \centering
            \begin{tikzpicture} [
            node distance = 4em,
            every text node part/.style={align=center},
            action/.style={rectangle, draw=black, rounded corners, minimum size=2mm},
            start chain=obj going right,
            object/.style={rectangle, draw=black, fill=Cerulean, rounded corners, minimum size=2mm, on chain=obj},
            pred_ag/.style={rectangle, draw=black, fill=BrickRed, rounded corners, minimum size=2mm},
            pred_av/.style={rectangle, draw=black, fill=ForestGreen, rounded corners, minimum size=2mm},
            pred_ug/.style={rectangle, draw=black, fill=Goldenrod, rounded corners, minimum size=2mm},
            param_1/.style={draw=orange},
            param_2/.style={draw=blue}
        ]   
            \node[action, fill=orange] (A_1) {$a_1(c_1, c_2)$};
            \node[action, fill=brown, right=5em] (A_2) {$a_2(c_2)$};
    
            \foreach \c [count=\i] in {1,2} {
                \ifnum\i=1
                    \node[object, yshift=-2.5em] (c_\c) {$c_\c$};
                \else
                    \node[object] (c_\c) {$c_\c$};
                \fi
            }
    
            \node[pred_ag, below=1em of c_1] (p_1) {$p_1(c_1)$};
            \node[pred_ug, below=1em of c_2] (p_2) {$p_2(c_2)$};
    
            \path[param_1] (A_1) edge (c_1);
            \path[param_2] (A_1) edge (c_2);
    
            \path[param_1] (A_2) edge (c_2);
    
            \path[param_1] (p_1) edge (c_1);
    
            \path[param_1] (p_2) edge (c_2);
            
        \end{tikzpicture}
        \caption{AOAG graph.}
        \label{fig:aoag}
    \end{subfigure}
    \hfill
        \begin{subfigure}[b]{0.48\linewidth}
        \centering
        \begin{tikzpicture} [
            node distance = 1em,
            every text node part/.style={align=center},
            start chain=var going right,
            var/.style={rectangle, draw=black, fill=Cerulean, rounded corners, minimum size=2mm, on chain=var},
            start chain=dom going right,
            dom_cur/.style={rectangle, draw=black, fill=ForestGreen, rounded corners, minimum size=2mm, on chain=dom},
            dom_goal/.style={rectangle, draw=black, fill=Goldenrod, rounded corners, minimum size=2mm, on chain=dom},
            dom_poss/.style={rectangle, draw=black, fill=violet, rounded corners, minimum size=2mm, on chain=dom},
            dom_agoal/.style={rectangle, draw=black, fill=BrickRed, rounded corners, minimum size=2mm, on chain=dom},
            action/.style={circle, draw=black},
            dom_edge/.style={draw=CadetBlue},
            pre/.style={draw=black},
            eff/.style={draw=blue}
        ]   
            \foreach \v in {1,2} \node[var] (v_\v) {$v_\v$};
    
            \begin{scope}[
                node distance = 0.5em
            ]
                \node[dom_agoal, above= of v_1] (d_1_1) {$T_1$};
                \node[dom_poss, left= of d_1_1] (d_1_0) {$F_1$};
    
                \path[dom_edge] (v_1) edge (d_1_0);
                \path[dom_edge] (v_1) edge (d_1_1);
    
                \node[dom_cur, above=2mm of v_2] (d_2_0) {$F_2$};
                \node[dom_goal] (d_2_1) {$T_2$};
    
                \path[dom_edge] (v_2) edge (d_2_0);
                \path[dom_edge] (v_2) edge (d_2_1);
    
            \end{scope}
    
            \node[action, fill=orange, above= 1em of d_1_1] (action_1) {$o_1$};
            \node[action, fill=brown, above=1em of d_2_0, xshift=1em] (action_2) {$o_2$};
    
            \path[pre] (d_1_0) edge (action_1);
            \path[pre] (d_2_1) edge (action_1);
            
            \path[pre] (d_2_0) edge (action_2);
    
            \path[eff] (d_1_1) edge (action_1);
            \path[eff] (d_2_0) edge (action_1);
            \path[eff] (d_2_1) edge (action_2);
    
            
        \end{tikzpicture}
        \caption{FLG graph.}
        \label{fig:flg}
    \end{subfigure}
    
    \caption{Graph representations for $\task_X$ in Table~\ref{tab:example_task}.}
    \label{fig:graphs}
\end{figure}

Fig.~\ref{fig:aoag} illustrates the AOAG graph of the example task defined in Table~\ref{tab:example_task}.

\subsubsection{FDR learning graph}\label{sec:flg}
We provide an alternative to the original FLG from~\citet{chen:etal:24:aaai}
that is made compatible with the WL algorithm  by integrating predicate and action schema information as node colours.
The graph's nodes represent the FDR task variables, values, and actions. Action nodes are labeled by their corresponding action schema,
whereas the label of a value node encodes the predicate it stems from (via the function $\rho$) and its ``goal status'', similarly as for the AOAG. Edges connect variables with their possible values, and these values to the actions that use them as preconditions or effects. Edges
are labelled with the type of edge they represent (i.e. variable-value, precondition, or effect edge).
Formally, the FLG of a task \task is the graph $\graphFLG (s) = \langle \nodes, \edges, \ncolours, \ecolours \rangle$ with:
\begin{itemize}
    \item $\nodes = \vars \cup \mathcal{D}_\task \cup \ops$
    \item $\edges = \evarval \cup \epre \cup \eeff$ with \\ 
        $\evarval = \bigcup_{v \in \vars} \{\langle v, d\rangle_\texttt{var:val} \ | \ d \in \dom{v} \}$, \\
        $\epre = \bigcup_{o \in O} \{\langle o, d\rangle_\texttt{pre} \ | \ \exists v \langle v, d\rangle \in \pre{o} \}$ and \\
        $\eeff = \bigcup_{o \in O} \{\langle o, d\rangle_\texttt{eff} \ | \ \exists v \langle v, d\rangle \in \eff{o} \}$
    \item $\ncolours: V \rightarrow \{\texttt{var}\} \cup (\{\texttt{av},\texttt{uv},\texttt{ag},\texttt{ug}\} \times \preds) \cup \schemas$ defined by $u \rightarrow$
        \begin{equation*}
          \begin{cases}
            \texttt{var}, & \text{if $u \in \vars$}.\\
            (\texttt{av}, \rho(d)), & \text{if $u = d, \ \exists v \ \langle v,d \rangle \in s \setminus s_*$}.\\
            (\texttt{uv}, \rho(d)), & \text{if $u = d, \ \exists v \ d \in \dom{v}, \langle v, d \rangle \not \in (s \ \cup \ s_*)$}.\\
            (\texttt{ag}, \rho(d)), & \text{if $u = d, \ \exists v \ \langle v,d \rangle \in s \ \cap \ s_*$}.\\
            (\texttt{ug}, \rho(d)), & \text{if $u = d, \ \exists v \ \langle v,d \rangle \in s_* \setminus s$}.\\
            sch(u), & \text{if $u \in O$}.\\
            \end{cases}
        \end{equation*}
    \item $\ecolours: E \rightarrow \{\texttt{var:varl},\texttt{pre},\texttt{eff}\}$ with $l(e) = \alpha$ for $e \in E_\alpha$
\end{itemize}

Fig.~\ref{fig:flg} illustrates the FLG graph of the example task defined in Table~\ref{tab:example_task}.

\subsubsection{Project to pattern abstraction}\label{sec:project}

Now that a planning state can be encoded as a labelled graph $\graph (s)$, we can adapt this same graph to a given pattern \pattern to capture the information specific to the state $s$ in the projected task \taskP.
We propose a generic masking method that projects $\graph (s)$ onto any chosen pattern \pattern.
Starting from the graph construction for either AOAG or FLG, we obtain the projected graph $\graph (s)^\pattern$ by masking (i.e., removing) all vertices and edges that are not relevant to the variables in \pattern.
For FLG, masking is applied to nodes that represent a variable $v$ and their domain $\dom{v}$ such that $v \not\in \pattern$.
Since AOAG is a lifted representation, we first ground the task using the \sasp translation~\cite{helmert:09}, which assigns each object and atom node to the corresponding \sasp variables.
We then mask all nodes 
associated with variables $v \not\in \pattern$. 
Edges are also masked from the graph whenever at least one of their incident nodes is masked.
However, action nodes are never masked, even when they become disconnected.
This is because the feature extraction step must produce an embedding for every ground action of the pattern abstraction, regardless of its connectivity in the projected subproblem. Consequently, $\graph (s)^\pattern$ may contain isolated action nodes.

\subsection{Action-centric Feature Extraction}\label{sec:action_wl}

To predict a cost partition for a set of patterns $\patternset$ and a state $s$, our framework must extract informative features for every action $o \in \ops$ and every pattern $\pattern \in \patternset$ from the $\graph(s)^\pattern$ projection.
Cost partitioning distributes the original cost of each action across the abstractions, subject to the constraint~(\ref{eqn:cost-partitioning-constraint}) and the quality of the resulting additive heuristic critically depends on how the distribution is performed.
An effective partition allocates a larger share of an action's cost to those abstractions where the action is structurally important, i.e., where it contributes to a strong heuristic estimate.
Therefore, the features must capture the structural role of $o$ within the pattern projection \taskP, for $\pattern \in \patternset$ at state $s$, enabling the subsequent model to infer high‑quality cost partitions.

\begin{algorithm}[tb]
\caption{Action-centric WL algorithm}
\label{alg:action_wl}
\textbf{Input}: Projected graph $\graph(s)^\pattern = \langle \nodes, \edges, \ncolours, \ecolours \rangle$; set of action \ops; number of iterations \iters.\\
\textbf{Output}: List of set of colours.
\begin{algorithmic}[1] 
\STATE $\ncolours^{0}(v) \leftarrow \ncolours(v), \forall v \in V$ \label{line:action_wl:init}\\ 
\FOR{$o \in \ops$}
\STATE $H^o \leftarrow \{o\} \cup \{u \mid u \in \neighboors(o) \} $\label{line:action_wl:hop}\\
\ENDFOR
\FOR{$j=1,\ldots,\iters$ \normalfont{\textbf{do for}} $v \in V$}
\STATE $\ncolours^{j}(v) \leftarrow \text{hash} \big(\ncolours^{j-1}(v), \{(\ncolours^{j-1}(u), \iota) \mid \iota \in \Sigma_E, u \in \neighboors_\iota (v)\} \big)$
\label{line:action_wl:update}
\ENDFOR
\STATE \textbf{return} {$\big\langle \bigcup_{j=0,\ldots,L}\{\ncolours^{j}(u) \mid u \in H^o \} \mid o \in \ops \big\rangle$} 
\label{line:action_wl:return}
\end{algorithmic}
\end{algorithm}

To compute per-node embeddings, we introduce a new Action-centric WL (AWL) in Alg.~\ref{alg:action_wl} based on WL procedure from Section~\ref{sec:graph_rep}.
A projected graph $\graph (s)^\pattern$ and the set of action \ops are taken as input along with the number of iterations \iters.
Line~\ref{line:action_wl:hop} collects a hop set $H^o$ for each action node $o \in \ops$ that contains $o$ together with its immediate neighbours.
This hop set helps to capture the local structural context of an action.
After the standard WL update on every node at Line~\ref{line:action_wl:update}, the algorithm returns a set of colours that appears in the hop set $H^o$ of each action $o$ across all \iters iterations.
We can represent each of those sets as a histogram feature vector described in Section~\ref{sec:graph_rep}.

For a state $s$, we apply this process to each $\pattern \in \patternset$ and arrange the resulting vectors into a three-dimensional feature array.
Let \odim the number of actions such that $\odim = |\ops|$, and \pdim the number of patterns such that $\pdim = |\patternset|$.
The array \Xdims{s} stacks the per‑action feature vector for every pattern-action pair in the order of the action and pattern enumeration (see Fig.~\ref{fig:overview}).
We use the same set of actions \ops for every \pattern, thus the array has a consistent shape across patterns.
The first dimension indexes the abstraction, the second ranges over the actions, and the third contains the AWL feature vector of the action node.
This array is the input to the learned partition predictor described next.

\subsection{Axial Self-Attention to predict Cost Partitions}
\label{sec:model_architecture}
Now that we have defined how to extract action‑centric features into the tensor $\X(s)$, we present the model that transforms these features into admissible cost partitions.

The feature array \Xdims{s} produced by the AWL‑based extraction encodes, for each abstraction and each ground action, a structural signature of how that action appears within the pattern projection at state $s$.
To turn this information into a cost partition, we need a model that can simultaneously reason about two types of context: the relative importance of actions inside a single abstraction, and the role an action plays across different abstractions.
Once the reasoning is done, the model needs to compute a cost weight for each action that reflects its significance across all abstractions.
To enforce admissibility, those weights must respect the
cost partition constraint~(\ref{eqn:cost-partitioning-constraint}), i.e., for every action the weights across abstractions must sum to at most its original cost.
We propose to achieve this with an axial self‑attention block~\cite{ho:etal:2019} followed by a multi‑layer perceptron (MLP) and a softmax output layer.

\subsubsection{Axial self-attention}
Self‑attention~\cite{vaswani:etal:17} computes, for each input embedding, a weighted sum over all embeddings, where the weights are derived via query–key similarity (e.g., softmax of scaled dot‑products).
This operation is inherently agnostic to sequence length and order because the attention pattern depends only on pairwise relationships, not on absolute positions or a fixed receptive field.
This design allows a model trained on short sequences to be deployed on longer ones without architectural modification, enabling length extrapolation.
Axial self-attention applies self‑attention along distinct axes of the input array, allowing the model to contextualize the action features from two complementary perspectives.

\begin{enumerate}
    \item \textbf{Action self‑attention.}
    Along the action axis, for each abstraction $i \in \{1, \dots, \pdim\}$ we treat the \odim action‑node embeddings $\X(s)_{i,\cdot,\cdot} \in \mathbb{R}^{\odim \times \vdim}$ as a sequence, and apply a standard self‑attention mechanism.
    This allows the model to compare actions within the same abstraction, answering: \emph{which actions in this pattern should receive a larger share of cost?}
    \item \textbf{Abstraction self‑attention.} Along the abstraction axis, for each action position $j \in \{1, \dots, \odim\}$ we slice the array as $\X(s)_{\cdot,j,\cdot} \in \mathbb{R}^{\pdim \times \vdim}$, and again apply self‑attention. This contextualizes each action by looking at how it is represented across different patterns, answering: \emph{in which kinds of abstractions is this action structurally important?}
\end{enumerate}

We apply action self-attention first and abstraction self-attention on its output.
The inputs of the two self-attention is passed through layer normalization.
Outputs goes through a linear projection and a dropout layer before being fused with their input by element-wise addition.
Final output is a contextualized array $\carray$ with the same spatial dimensions as $\X(s)$.

\subsubsection{Score matrix}
The contextualized array $\carray$ is passed through  then fed into a point‑wise MLP applied independently to every $(i,j)$ position.
This MLP consist of two hidden layers with ReLU activations, followed by a dropout layer and a linear projection that reduces the feature dimension from \vdim to $1$.
For each abstraction $i$ and action $j$, the MLP outputs a scalar score $S_ {i,j}$ resulting in a score matrix $S^{\pdim \times \odim}$. A higher score indicates that the action is particularly salient for the abstraction, and therefore a larger portion of the original action cost should be allocated.

\subsubsection{Admissible cost partition}\label{sec:admissible_lcp}
Finally, we convert the scores into valid cost‑partition weights.
For each action $j$, we interpret the vector $S_{\cdot,j}$ as the relative importance of the \pdim patterns for that action at state $s$.
To turn these scores into a proper allocation, we apply the softmax function over the pattern dimension:
\begin{equation*}\label{eq:softmax}
    \alpha_{i,j} = \frac{e^{S_{i,j}}}{\sum_{i'=1}^\pdim e^{S_{i',j}}}
\end{equation*}

The softmax properties guarantee $\sum_{i=1}^\pdim \alpha_{i,j} = 1, \forall j \in 1, \dots, \odim$ which is essential for admissibility.
These coefficients can be interpreted as the fraction of the original cost of action $o_j \in \ops$ that should be assigned to pattern $i$.
The predicted cost is then obtained by multiplying the original action cost by the coefficient $C_i(o_j) = \alpha_{i,j} \cdot \cost(o_j)$.
Since the coefficients sum to one, we immediately satisfies the cost partition constraint~(\ref{eqn:cost-partitioning-constraint})
as an equality.

\subsubsection{Loss function}
During training, we have access to the optimal cost partition for $s$, computed offline by solving OCP LP.
This gives a target allocation matrix $\alpha^* \in [0,1]^{\pdim \times \odim}$, where $\alpha^*_{i,j}$ is the fraction of $\cost(o_j)$ that the optimal partition assigns to pattern $i$.

We train the model to match this optimal behaviour by minimising the Kullback–Leibler (KL) divergence~\cite{kullback:leibler:1951} between the target and predicted distributions, summed over all actions:
\begin{equation*}
    \loss(s) = \sum_{j=1}^{\odim} \sum_{i=1}^{\pdim} 
\alpha^*_{i,j} \log\frac{\alpha^*_{i,j}}{\alpha_{i,j}}
\end{equation*}
Minimising $\loss(s)$ encourages the network to allocate cost in the same proportion as the optimal partition.

\subsubsection{Training loop overview}
We train the model offline on synthetic planning tasks, using fixed training and validation splits where the validation tasks are larger than those used for training.
For each instance, a sampler uses random walks from the initial state to draw valid states and computes $\alpha^*_{i,j}$ via LP.
We implement a non-negative OCP LP~\cite{pommerening:etal:14} with a modified cost partition constraint~(\ref{eqn:cost-partitioning-constraint}):
\begin{equation*}
    \sum_{\pattern \in \patternset} \cost^\pattern (o) = \cost (o), \forall o \in O
\end{equation*}
to extract weights compatible with a softmax output.
For each mini‑batch of states, we compute the loss and backpropagate through the neural network. AWL feature extraction is not updated. 
After each epoch, we evaluate the loss on the validation set.

\section{Experiments}
In this section, we empirically evaluate our framework for learning domain-specific admissible cost partition heuristics on a subset of the benchmarks from the optimal track of the International Planning Competition (IPC). 
We train on optimal cost weights from sampled states of synthetic tasks described in Table~\ref{tab:synth_instances}.
We generate samples for each split using Scorpion~\cite{seipp:etal:20} with 500 maximum samples per task and 30min cutoff time.
We use the set of all projections up to 2 variables and limited to interesting patterns~\cite{pommerening:etal:13} with at least one goal variable.

We consider a combination of two hyperparameters in our configurations: graph representation $R \in \{\FLG, \AOAG\}$ and AWL iterations $L \in \{1,2\}$.
We train our model with batch size 4, initial learning rate of $10^{-5}$ and AdamW~\cite{loshchilov:hutter:19} optimizer with weight decay of 0.01 for a maximum of 100 epochs.
We apply early stopping if the validation loss does not improve by at least $10^{-4}$ for 15 epochs.
The learning rate also decreases by a factor of 10 if the minimal validation loss did not decrease in the last 5 epochs.
Hidden layers in point-wise MLP compress the embedding dimension $\vdim$ by a factor of 0.4 before restoring the original size right before the linear projection.
All other learning layers have a dimension equal to $\vdim$.
The dropout rate is 0.1 for the scalar dot-product attention mechanism in axial attention, 0.2 for dropout layers after both axial attention and 0.3 before the MLP.
Models are trained using PyTorch 2.11 on a cluster with 4 Intel Xeon Gold 6448Y cores, 62Go RAM and $\frac{3}{8}$th of the computing power of an NVidia H100 SXM5 with 40GB GPU memory.

\begin{table}[t]
    \centering

    \begin{tabular}{
        lllll
    }
        \toprule
        \textbf{Domain} & \textbf{Train} & \textbf{\#} & \textbf{Validation} & \textbf{\#} 
        \\
        \midrule
        blocks    & [4, 9]  & 27 & [10] & 6 
        \\
        ferry     & [2, 8]  & 25 & [9]  & 4 
        \\
        logistics & [2, 6]  & 30 & [7]  & 6 
        \\
        miconic   & [3, 11] & 32 & [12] & 4 
        \\
        spanner   & [2, 9]  & 40 & [10] & 5 
        \\
        visitall  & [2, 7]  & 35 & [8]  & 6 
        \\
        \bottomrule
    \end{tabular}
    \caption{Synthetic problem splits with corresponding numbers of tasks per domain.}
     \label{tab:synth_instances}
\end{table}

For final evaluation, we build an additive heuristic $\hLCP (s)$ on top of the predicted costs $C$ such that $\hLCP (s) = \sum_{\pattern_i \in \patternset} h^{P_i}(s, C_i)$.
We write \hLCPRL{R}{L} for the learned heuristic on graph representation $R$ with $L$ iterations.
Our heuristic predicts a new cost partition at every evaluated state and only uses this new partition to compute the heuristic value. 
The learned heuristic is integrated into the Fast Downward (FD) planning system~\cite{helmert:06} as a separate heuristic evaluator using the LibTorch C++ CUDA library.
We benchmark on the standard IPC optimal track instances for the same domains as training.
Since Ferry and Spanner are not part of the optimal track, we use the easy testing task from the 2023 IPC learning track.
We use three cost partitioning heuristics implemented in Scorpion as baselines: Greedy zero-one (\hGZOCP), Non-negative online saturated (\hSCP) and Non-negative optimal (\hOCP).
\hGZOCP and \hSCP both use a greedy ordering of the patterns, \hSCP computes a new cost partition at every evaluated state.
All methods are evaluated using A* and a timeout of 1800 seconds.
Baselines are run on a cluster with single Intel Xeon Gold 6448Y core and 32Go RAM.
Our learned heuristics also use $\frac{3}{8}$th of the computing power of an NVidia H100 SXM5 with 40GB GPU memory on the same cluster.
In our current implementation, the GPU is only used to speedup the learning model prediction of our framework.

\begin{table}[t]
\centering
\footnotesize
\setlength{\tabcolsep}{3pt}
\begin{tabular}{lccccccc}
\toprule

& blocks
& ferry
& logistics
& miconic
& spanner
& visitall
& \textbf{Total} \\ 
    & (35) 
    & (30) 
    & (63) 
    & (150) 
    & (30) 
    & (40) 
    & (348) \\
\midrule
\midrule
\hGZOCP 
    & \cellcolor{bestgray} \textbf{28} 
    & \cellcolor{secondgray} 13 
    & \cellcolor{secondgray} 27 
    & \cellcolor{secondgray} 70 
    & \cellcolor{bestgray} \textbf{30}
    & \cellcolor{secondgray} 30 
    & 198 \\

\hSCP 
    & \cellcolor{bestgray} \textbf{28} 
    & \cellcolor{bestgray} \textbf{22} 
    & \cellcolor{thirdgray} 26 
    & \cellcolor{bestgray} \textbf{146} 
    & \cellcolor{bestgray} \textbf{30}
    & \cellcolor{bestgray} \textbf{33} 
    & \textbf{285} \\

\hOCP 
    & 17 
    & 11 
    & \cellcolor{bestgray} \textbf{33} 
    & \cellcolor{thirdgray} 50
    & \cellcolor{bestgray} \textbf{30}
    & 16 
    & 157 \\
\midrule
\hLCPRL{\FLG}{1} 
    & \cellcolor{secondgray} 19 
    & 11 
    & 17
    & 40 
    & \cellcolor{bestgray} \textbf{30}
    & \cellcolor{thirdgray} 20 
    & 137 \\

\hLCPRL{\FLG}{2} 
    & \cellcolor{thirdgray} 18 
    & 11 
    & 18
    & 40 
    & \cellcolor{bestgray} \textbf{30}
    & \cellcolor{thirdgray} 20 
    & 137 \\

\hLCPRL{\AOAG}{1} 
    & 17
    & \cellcolor{thirdgray} 12 
    & 15
    & 41 
    & \cellcolor{bestgray} \textbf{30}
    & 18 
    & 133 \\

\hLCPRL{\AOAG}{2} 
    & \cellcolor{thirdgray} 18 
    & \cellcolor{thirdgray} 12
    & 18
    & 42 
    & \cellcolor{secondgray} 28
    & 16 
    & 134 \\
\bottomrule
\end{tabular}
\caption{Coverage results per domain. Numbers in brackets shows the total number of instances per column. Best results among admissible configurations are highlighted in bold.}
\label{tab:coverage}
\end{table}

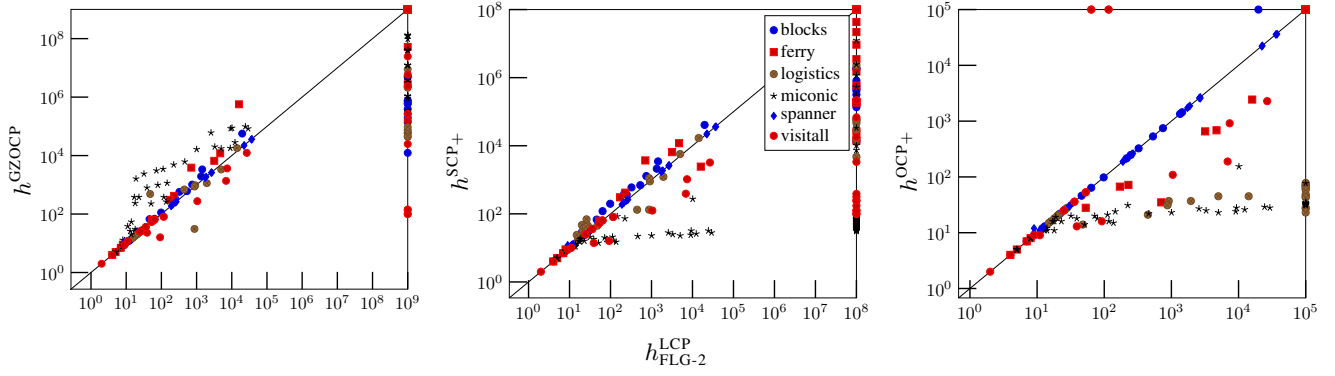
\begin{figure*}
    \centering
    \begin{tikzpicture}    
     \node (fig1){
        \begin{tikzpicture}[scale=0.65]
            \begin{axis}[height=8.00in, legend cell align=left, width=8.00in, xmax=1000000000, xmode=log, ymax=1000000000, ymode=log, ticklabel style={black},
            normalsize,
            every major tick/.append style={thick, major tick length=10pt, black},
            every minor tick/.append style={draw=none},
            xtick pos=left,ytick pos=left]
            \addplot+[only marks] coordinates {
            (1000000000, 709673) (1000000000, 139017) (1000000000, 584873) (1000000000, 622226) (1000000000, 392636) (1000000000, 199246) (1000000000, 776866) (1000000000, 12367) (1000000000, 1000000000) (1000000000, 1000000000) (1000000000, 4827889) (1000000000, 5172506) (1000000000, 1000000000) (1000000000, 1000000000) (1000000000, 1000000000) (1000000000, 1000000000) (1000000000, 1000000000) (7, 7) (13, 13) (9, 9) (22, 22) (21, 25) (46, 67) (20, 25) (12, 12) (325, 575) (98, 113) (1345, 1948) (248, 331) (216, 337) (1444, 3367) (64, 69) (19755, 56442) (532, 606) (748, 1018)
            };
            \addplot+[only marks] coordinates {
            (4, 4) (4698, 11945) (1000000000, 2757778) (706, 3867) (1000000000, 52138772) (1000000000, 1000000000) (1000000000, 1000000000) (1000000000, 1000000000) (1000000000, 1000000000) (1000000000, 1000000000) (1000000000, 1000000000) (5, 5) (1000000000, 1000000000) (1000000000, 1000000000) (1000000000, 1000000000) (1000000000, 1000000000) (1000000000, 1000000000) (1000000000, 1000000000) (1000000000, 1000000000) (1000000000, 1000000000) (1000000000, 1000000000) (1000000000, 1000000000) (7, 7) (1000000000, 1000000000) (8, 9) (53, 63) (229, 419) (171, 306) (3179, 6625) (15964, 567708)
            };
            \addplot+[only marks] coordinates {
            (1000000000, 43530) (1000000000, 61921) (1000000000, 58734) (1000000000, 2092420) (1000000000, 46316) (1000000000, 8955958) (1000000000, 1000000000) (1000000000, 1000000000) (1000000000, 1000000000) (1000000000, 1000000000) (1000000000, 1000000000) (1000000000, 1000000000) (21, 21) (20, 20) (16, 16) (28, 28) (18, 18) (9, 9) (26, 26) (15, 15) (26, 26) (25, 25) (914, 962) (14147, 17925) (877, 874) (5003, 3352) (1952, 1139) (873, 31) (1000000000, 74512) (1000000000, 1000000000) (1000000000, 1000000000) (1000000000, 1000000000) (1000000000, 94744) (1000000000, 1000000000) (1000000000, 1000000000) (1000000000, 1000000000) (1000000000, 1000000000) (1000000000, 1000000000) (1000000000, 1000000000) (1000000000, 1000000000) (1000000000, 1000000000) (1000000000, 1000000000) (1000000000, 1000000000) (1000000000, 1000000000) (1000000000, 1000000000) (1000000000, 1000000000) (1000000000, 1000000000) (1000000000, 1000000000) (1000000000, 1000000000) (1000000000, 1000000000) (1000000000, 1000000000) (1000000000, 1000000000) (1000000000, 1000000000) (1000000000, 1000000000) (1000000000, 1000000000) (1000000000, 1000000000) (1000000000, 1000000000) (1000000000, 1000000000) (48, 485) (446, 693) (1000000000, 265447) (1000000000, 1000000000) (1000000000, 1000000000)
            };
            \addplot+[only marks] coordinates {
            (5, 5) (8, 13) (1000000000, 1000000000) (1000000000, 1000000000) (1000000000, 1000000000) (1000000000, 1000000000) (1000000000, 1000000000) (1000000000, 1000000000) (1000000000, 1000000000) (1000000000, 1000000000) (1000000000, 1000000000) (1000000000, 1000000000) (11, 38) (1000000000, 1000000000) (1000000000, 1000000000) (1000000000, 1000000000) (1000000000, 1000000000) (1000000000, 1000000000) (1000000000, 1000000000) (1000000000, 1000000000) (1000000000, 1000000000) (1000000000, 1000000000) (1000000000, 1000000000) (14, 54) (1000000000, 1000000000) (1000000000, 1000000000) (1000000000, 1000000000) (1000000000, 1000000000) (1000000000, 1000000000) (1000000000, 1000000000) (1000000000, 1000000000) (1000000000, 1000000000) (1000000000, 1000000000) (1000000000, 1000000000) (14, 20) (1000000000, 1000000000) (1000000000, 1000000000) (1000000000, 1000000000) (1000000000, 1000000000) (1000000000, 1000000000) (1000000000, 1000000000) (1000000000, 1000000000) (1000000000, 1000000000) (1000000000, 1000000000) (1000000000, 1000000000) (18, 30) (1000000000, 1000000000) (1000000000, 1000000000) (1000000000, 1000000000) (1000000000, 1000000000) (1000000000, 1000000000) (1000000000, 1000000000) (1000000000, 1000000000) (1000000000, 1000000000) (1000000000, 1000000000) (1000000000, 1000000000) (13, 26) (1000000000, 1000000000) (137, 263) (51, 227) (17, 377) (23, 377) (4, 4) (117, 377) (79, 984) (146, 1147) (18, 234) (39, 796) (19, 1615) (32, 2386) (72, 3066) (112, 3449) (224, 4877) (5, 5) (461, 6176) (9093, 17936) (3290, 19985) (5009, 17400) (1004, 16398) (12347, 28169) (29276, 83519) (10133, 87936) (2574, 60544) (24424, 98489) (5, 5) (9308, 84212) (1000000000, 328074) (1000000000, 304181) (1000000000, 306567) (1000000000, 346866) (1000000000, 300348) (1000000000, 1060470) (1000000000, 1072121) (1000000000, 915581) (1000000000, 1101423) (5, 5) (1000000000, 1095140) (1000000000, 3733623) (1000000000, 3005031) (1000000000, 3794613) (1000000000, 3841965) (1000000000, 3154874) (1000000000, 12368301) (1000000000, 12177854) (1000000000, 12473923) (1000000000, 12091337) (8, 8) (1000000000, 12420025) (1000000000, 41105768) (1000000000, 39652348) (1000000000, 41379376) (1000000000, 40672109) (1000000000, 37877961) (1000000000, 128972508) (1000000000, 132121434) (1000000000, 133796652) (1000000000, 96173613) (8, 8) (1000000000, 123826377) (1000000000, 1000000000) (1000000000, 1000000000) (1000000000, 1000000000) (1000000000, 1000000000) (1000000000, 1000000000) (1000000000, 1000000000) (1000000000, 1000000000) (1000000000, 1000000000) (1000000000, 1000000000) (8, 8) (1000000000, 1000000000) (1000000000, 1000000000) (1000000000, 1000000000) (1000000000, 1000000000) (1000000000, 1000000000) (1000000000, 1000000000) (1000000000, 1000000000) (1000000000, 1000000000) (1000000000, 1000000000) (1000000000, 1000000000) (8, 8) (1000000000, 1000000000) (1000000000, 1000000000) (1000000000, 1000000000) (1000000000, 1000000000) (1000000000, 1000000000) (1000000000, 1000000000) (1000000000, 1000000000) (1000000000, 1000000000) (1000000000, 1000000000) (1000000000, 1000000000)
            };
            \addplot+[only marks] coordinates {
            (9, 12) (9, 9) (11, 11) (11, 11) (9, 10) (12, 12) (30, 30) (30, 30) (31, 31) (34, 34) (36, 36) (32, 32) (211, 209) (188, 188) (192, 190) (259, 259) (269, 263) (244, 244) (1888, 1873) (1729, 1729) (1801, 1801) (2593, 2593) (2709, 2573) (2686, 2686) (22352, 22064) (22154, 21976) (22686, 22495) (37652, 36320) (35869, 35684) (36373, 35979)
            };
            \addplot+[only marks] coordinates {
            (4, 4) (2, 2) (9, 9) (11, 12) (92, 16) (39, 23) (25, 25) (1047, 276) (36, 36) (6874, 1366) (53, 54) (1000000000, 161654) (64, 64) (1000000000, 24591210) (116, 81) (1000000000, 1000000000) (1000000000, 100) (1000000000, 1000000000) (1000000000, 142) (1000000000, 1000000000) (1000000000, 1000000000) (7373, 3639) (26661, 12369) (1000000000, 268571) (1000000000, 5898771) (1000000000, 1000000000) (1000000000, 100) (1000000000, 142) (1000000000, 2389627) (1000000000, 24860) (1000000000, 1000000000) (1000000000, 1000000000) (1000000000, 1000000000) (1000000000, 1000000000) (1000000000, 1000000000) (25, 25) (36, 36) (53, 54) (64, 64) (116, 81)
            };
            \draw[color=black] (axis cs:1e-70,1e-70) -- (axis cs:1e70,1e70);
            \end{axis}
        \end{tikzpicture}
     };
     \node (fig2) [right=1em of fig1]{
        \begin{tikzpicture}[scale=0.67]
            \begin{axis}[height=8.00in, legend cell align=left, width=8.00in, xmax=100000000, xmode=log, ymax=100000000, ymode=log, ticklabel style={black},
            normalsize,
            every major tick/.append style={thick, major tick length=10pt, black},
            every minor tick/.append style={draw=none},
            xtick pos=left,ytick pos=left]
            \addplot+[only marks] coordinates {
            (100000000, 816850) (100000000, 164284) (100000000, 472901) (100000000, 316347) (100000000, 408720) (100000000, 135934) (100000000, 574814) (100000000, 13480) (100000000, 100000000) (100000000, 100000000) (100000000, 1921487) (100000000, 1797327) (100000000, 100000000) (100000000, 100000000) (100000000, 100000000) (100000000, 100000000) (100000000, 100000000) (7, 7) (13, 13) (9, 9) (22, 23) (21, 28) (46, 68) (20, 29) (12, 12) (325, 598) (98, 198) (1345, 2115) (248, 369) (216, 394) (1444, 3472) (64, 120) (19755, 40891) (532, 695) (748, 1288)
            };
            \addlegendentry{blocks}
            \addplot+[only marks] coordinates {
            (4, 4) (4698, 11884) (100000000, 14119) (706, 3707) (100000000, 196349) (100000000, 187189) (100000000, 580215) (100000000, 26304) (100000000, 9333888) (100000000, 3485838) (100000000, 187419) (5, 5) (100000000, 100000000) (100000000, 21863222) (100000000, 100000000) (100000000, 100000000) (100000000, 100000000) (100000000, 100000000) (100000000, 1514129) (100000000, 43182074) (100000000, 100000000) (100000000, 100000000) (7, 7) (100000000, 100000000) (8, 9) (53, 63) (229, 419) (171, 306) (3179, 6505) (15964, 2437)
            };
            \addlegendentry{ferry}
            \addplot+[only marks] coordinates {
            (100000000, 46018) (100000000, 60280) (100000000, 56986) (100000000, 2306461) (100000000, 54772) (100000000, 100000000) (100000000, 100000000) (100000000, 100000000) (100000000, 100000000) (100000000, 100000000) (100000000, 100000000) (100000000, 100000000) (21, 32) (20, 47) (16, 17) (28, 45) (18, 22) (9, 9) (26, 69) (15, 24) (26, 31) (25, 41) (914, 903) (14147, 16891) (877, 1114) (5003, 5715) (1952, 1222) (873, 133) (100000000, 4737) (100000000, 100000000) (100000000, 100000000) (100000000, 100000000) (100000000, 15248) (100000000, 100000000) (100000000, 100000000) (100000000, 100000000) (100000000, 100000000) (100000000, 100000000) (100000000, 100000000) (100000000, 100000000) (100000000, 100000000) (100000000, 100000000) (100000000, 100000000) (100000000, 100000000) (100000000, 100000000) (100000000, 100000000) (100000000, 100000000) (100000000, 100000000) (100000000, 100000000) (100000000, 100000000) (100000000, 100000000) (100000000, 100000000) (100000000, 100000000) (100000000, 100000000) (100000000, 100000000) (100000000, 100000000) (100000000, 100000000) (100000000, 100000000) (48, 45) (446, 132) (100000000, 28240) (100000000, 100000000) (100000000, 100000000)
            };
            \addlegendentry{logistics}
            \addplot+[only marks] coordinates {
            (5, 5) (8, 8) (100000000, 66) (100000000, 71) (100000000, 71) (100000000, 70) (100000000, 71) (100000000, 72) (100000000, 76) (100000000, 71) (100000000, 75) (100000000, 77) (11, 11) (100000000, 73) (100000000, 82) (100000000, 12472746) (100000000, 75) (100000000, 525782) (100000000, 75) (100000000, 80) (100000000, 81) (100000000, 80) (100000000, 81) (14, 13) (100000000, 81) (100000000, 85) (100000000, 90) (100000000, 2428308) (100000000, 87) (100000000, 87) (100000000, 85) (100000000, 100000000) (100000000, 92) (100000000, 91) (14, 11) (100000000, 84) (100000000, 100000000) (100000000, 92) (100000000, 91) (100000000, 92) (100000000, 91) (100000000, 97) (100000000, 100000000) (100000000, 101) (100000000, 89) (18, 15) (100000000, 94) (100000000, 111) (100000000, 96) (100000000, 100) (100000000, 100) (100000000, 101) (100000000, 98) (100000000, 101) (100000000, 101) (100000000, 100000000) (13, 11) (100000000, 106) (137, 15) (51, 14) (17, 16) (23, 16) (4, 4) (117, 16) (79, 18) (146, 84) (18, 18) (39, 18) (19, 19) (32, 21) (72, 20) (112, 21) (224, 73) (5, 5) (461, 22) (9093, 24) (3290, 25) (5009, 23) (1004, 23) (12347, 26) (29276, 28) (10133, 273) (2574, 28) (24424, 32) (5, 5) (9308, 32) (100000000, 32) (100000000, 32) (100000000, 31) (100000000, 33) (100000000, 135) (100000000, 35) (100000000, 34) (100000000, 34) (100000000, 37) (5, 5) (100000000, 34) (100000000, 40) (100000000, 36) (100000000, 40) (100000000, 10168) (100000000, 37) (100000000, 41) (100000000, 43) (100000000, 41) (100000000, 40) (8, 8) (100000000, 42) (100000000, 46) (100000000, 7420) (100000000, 48) (100000000, 45) (100000000, 42) (100000000, 46) (100000000, 49) (100000000, 51) (100000000, 45) (8, 8) (100000000, 47) (100000000, 50) (100000000, 51) (100000000, 51) (100000000, 53) (100000000, 48) (100000000, 54) (100000000, 64) (100000000, 56) (100000000, 61) (8, 8) (100000000, 55) (100000000, 60) (100000000, 55) (100000000, 56) (100000000, 62) (100000000, 33664) (100000000, 63) (100000000, 61) (100000000, 62) (100000000, 61) (8, 8) (100000000, 58) (100000000, 69) (100000000, 68) (100000000, 1176040) (100000000, 71) (100000000, 64) (100000000, 296622) (100000000, 68) (100000000, 71) (100000000, 66)
            };
            \addlegendentry{miconic}
            \addplot+[only marks] coordinates {
            (9, 12) (9, 9) (11, 11) (11, 11) (9, 10) (12, 12) (30, 30) (30, 30) (31, 31) (34, 34) (36, 36) (32, 32) (211, 209) (188, 188) (192, 190) (259, 259) (269, 263) (244, 244) (1888, 1873) (1729, 1729) (1801, 1801) (2593, 2593) (2709, 2573) (2686, 2686) (22352, 22064) (22154, 21976) (22686, 22495) (37652, 36320) (35869, 35684) (36373, 35979)
            };
            \addlegendentry{spanner}
            \addplot+[only marks] coordinates {
            (4, 4) (2, 2) (9, 9) (11, 10) (92, 16) (39, 14) (25, 25) (1047, 125) (36, 36) (6874, 391) (53, 53) (100000000, 3361) (64, 64) (100000000, 217864) (116, 81) (100000000, 100000000) (100000000, 100) (100000000, 100000000) (100000000, 125) (100000000, 100000000) (100000000, 100000000) (7373, 1040) (26661, 3220) (100000000, 18291) (100000000, 68375) (100000000, 100000000) (100000000, 100) (100000000, 125) (100000000, 240) (100000000, 174) (100000000, 100000000) (100000000, 251) (100000000, 100000000) (100000000, 316) (100000000, 386) (25, 25) (36, 36) (53, 53) (64, 64) (116, 81)
            };
            \addlegendentry{visitall}
            \draw[color=black] (axis cs:1e-70,1e-70) -- (axis cs:1e70,1e70);
            \end{axis}
        \end{tikzpicture}
     };
     \node (fig3) [right=1em of fig2]{
        \begin{tikzpicture}[scale=0.67]
            \begin{axis}[height=8.00in, width=8.00in, xmax=100000, xmode=log, ymax=100000, ymode=log, ticklabel style={black},
            normalsize,
            every major tick/.append style={thick, major tick length=10pt, black},
            every minor tick/.append style={draw=none},
            xtick pos=left,ytick pos=left]
            \addplot+[only marks] coordinates {
            (100000, 100000) (100000, 100000) (100000, 100000) (100000, 100000) (100000, 100000) (100000, 100000) (100000, 100000) (100000, 100000) (100000, 100000) (100000, 100000) (100000, 100000) (100000, 100000) (100000, 100000) (100000, 100000) (100000, 100000) (100000, 100000) (100000, 100000) (7, 7) (13, 13) (9, 9) (22, 22) (21, 21) (46, 46) (20, 20) (12, 12) (325, 325) (98, 98) (1345, 1345) (248, 248) (216, 216) (1444, 1444) (64, 64) (19755, 100000) (532, 532) (748, 748)
            };
            \addplot+[only marks] coordinates {
            (4, 4) (4698, 686) (100000, 100000) (706, 35) (100000, 100000) (100000, 100000) (100000, 100000) (100000, 100000) (100000, 100000) (100000, 100000) (100000, 100000) (5, 5) (100000, 100000) (100000, 100000) (100000, 100000) (100000, 100000) (100000, 100000) (100000, 100000) (100000, 100000) (100000, 100000) (100000, 100000) (100000, 100000) (7, 7) (100000, 100000) (8, 8) (53, 28) (229, 72) (171, 67) (3179, 655) (15964, 2437)
            };
            \addplot+[only marks] coordinates {
            (100000, 46) (100000, 43) (100000, 49) (100000, 61) (100000, 43) (100000, 69) (100000, 74) (100000, 65) (100000, 59) (100000, 72) (100000, 78) (100000, 68) (21, 21) (20, 20) (16, 16) (28, 28) (18, 18) (9, 9) (26, 26) (15, 15) (26, 26) (25, 25) (914, 37) (14147, 45) (877, 32) (5003, 45) (1952, 37) (873, 31) (100000, 27) (100000, 100000) (100000, 100000) (100000, 100000) (100000, 23) (100000, 100000) (100000, 100000) (100000, 100000) (100000, 100000) (100000, 100000) (100000, 100000) (100000, 100000) (100000, 100000) (100000, 100000) (100000, 100000) (100000, 100000) (100000, 100000) (100000, 100000) (100000, 100000) (100000, 100000) (100000, 100000) (100000, 100000) (100000, 100000) (100000, 100000) (100000, 100000) (100000, 100000) (100000, 100000) (100000, 100000) (100000, 100000) (100000, 100000) (48, 14) (446, 21) (100000, 28) (100000, 100000) (100000, 100000)
            };
            \addplot+[only marks] coordinates {
            (5, 5) (8, 8) (100000, 100000) (100000, 100000) (100000, 100000) (100000, 100000) (100000, 100000) (100000, 100000) (100000, 100000) (100000, 100000) (100000, 100000) (100000, 100000) (11, 11) (100000, 100000) (100000, 100000) (100000, 100000) (100000, 100000) (100000, 100000) (100000, 100000) (100000, 100000) (100000, 100000) (100000, 100000) (100000, 100000) (14, 12) (100000, 100000) (100000, 100000) (100000, 100000) (100000, 100000) (100000, 100000) (100000, 100000) (100000, 100000) (100000, 100000) (100000, 100000) (100000, 100000) (14, 11) (100000, 100000) (100000, 100000) (100000, 100000) (100000, 100000) (100000, 100000) (100000, 100000) (100000, 100000) (100000, 100000) (100000, 100000) (100000, 100000) (18, 11) (100000, 100000) (100000, 100000) (100000, 100000) (100000, 100000) (100000, 100000) (100000, 100000) (100000, 100000) (100000, 100000) (100000, 100000) (100000, 100000) (13, 11) (100000, 100000) (137, 15) (51, 14) (17, 16) (23, 16) (4, 4) (117, 16) (79, 18) (146, 24) (18, 16) (39, 18) (19, 19) (32, 20) (72, 20) (112, 21) (224, 31) (5, 5) (461, 22) (9093, 24) (3290, 25) (5009, 23) (1004, 23) (12347, 26) (29276, 28) (10133, 154) (2574, 27) (24424, 29) (5, 5) (9308, 28) (100000, 32) (100000, 31) (100000, 31) (100000, 33) (100000, 77) (100000, 34) (100000, 33) (100000, 33) (100000, 35) (5, 5) (100000, 34) (100000, 100000) (100000, 100000) (100000, 100000) (100000, 100000) (100000, 100000) (100000, 100000) (100000, 100000) (100000, 100000) (100000, 100000) (8, 8) (100000, 100000) (100000, 100000) (100000, 100000) (100000, 100000) (100000, 100000) (100000, 100000) (100000, 100000) (100000, 100000) (100000, 100000) (100000, 100000) (8, 8) (100000, 100000) (100000, 100000) (100000, 100000) (100000, 100000) (100000, 100000) (100000, 100000) (100000, 100000) (100000, 100000) (100000, 100000) (100000, 100000) (8, 8) (100000, 100000) (100000, 100000) (100000, 100000) (100000, 100000) (100000, 100000) (100000, 100000) (100000, 100000) (100000, 100000) (100000, 100000) (100000, 100000) (8, 8) (100000, 100000) (100000, 100000) (100000, 100000) (100000, 100000) (100000, 100000) (100000, 100000) (100000, 100000) (100000, 100000) (100000, 100000) (100000, 100000)
            };
            \addplot+[only marks] coordinates {
            (9, 12) (9, 9) (11, 11) (11, 11) (9, 10) (12, 12) (30, 30) (30, 30) (31, 31) (34, 34) (36, 36) (32, 32) (211, 209) (188, 188) (192, 190) (259, 259) (269, 263) (244, 244) (1888, 1873) (1729, 1729) (1801, 1801) (2593, 2593) (2709, 2573) (2686, 2686) (22352, 22064) (22154, 21976) (22686, 22495) (37652, 36320) (35869, 35684) (36373, 35979)
            };
            \addplot+[only marks] coordinates {
            (4, 4) (2, 2) (9, 9) (11, 9) (92, 16) (39, 13) (25, 25) (1047, 109) (36, 36) (6874, 189) (53, 53) (100000, 100000) (64, 100000) (100000, 100000) (116, 100000) (100000, 100000) (100000, 100000) (100000, 100000) (100000, 100000) (100000, 100000) (100000, 100000) (7373, 915) (26661, 2283) (100000, 100000) (100000, 100000) (100000, 100000) (100000, 100000) (100000, 100000) (100000, 100000) (100000, 100000) (100000, 100000) (100000, 100000) (100000, 100000) (100000, 100000) (100000, 100000) (25, 25) (36, 36) (53, 53) (64, 100000) (116, 100000)
            };
            \draw[color=black] (axis cs:1e-70,1e-70) -- (axis cs:1e70,1e70);
            \end{axis}
        \end{tikzpicture}
     };
     \path ([xshift=2mm]fig1.south west)  -- ([xshift=2mm]fig1.north west) node[midway,above,sloped]{\small \hGZOCP};
     \path ([xshift=2mm]fig2.south west)  -- ([xshift=2mm]fig2.north west) node[midway,above,sloped]{\small \hSCP};
     \path ([xshift=2mm]fig3.south west)  -- ([xshift=2mm]fig3.north west) node[midway,above,sloped]{\small \hOCP};
     \path (fig1.south)  -- (fig3.south) node[midway, below]{\small \hLCPRL{\FLG}{2}};
     \end{tikzpicture}
    \caption{Number of expanded nodes of \hGZOCP, \hSCP, \hOCP and \hLCPRL{\FLG}{2}. On each plots, unsolved task by one planner has their respective metric set to the axis limit. Points on the top left triangle favour \hLCPRL{\FLG}{2} while points on the bottom right triangle favour baselines.}
    \label{fig:exp_nodes}
\end{figure*}

Table~\ref{tab:coverage} reports the number of tasks solved within the resource limits and Figure~\ref{fig:exp_nodes} the number of nodes expanded by \hLCPRL{\FLG}{2} vs the baselines.
In the remainder of this section, we discuss our results and answer the following questions.

\subsubsection{How well do our learned heuristics perform?}
Full coverage results are given in Table~\ref{tab:coverage}.
On every instance solved by both a learned heuristic and an admissible baseline, the resulting plan has the same cost, empirically confirming the admissibility guarantee of Section~\ref{sec:admissible_lcp}.
Our learned configurations achieve competitive coverage with \hOCP on most domains; the exceptions are logistics and miconic, where they fall behind.
On all domains, however, the learned heuristics are outperformed by the faster \hSCP and \hGZOCP baselines because their higher per‑state evaluation cost outweighs the gain in node expansions.
We attribute the poor performance on logistics and miconic, at least in part, to the absence of static atoms from our graph representation, as both domains rely heavily on such static relations.
Our graphs may therefore be insufficiently expressive in these settings, preventing the model from inferring appropriate cost partitions and leading to weak heuristic values on larger problems.

\subsubsection{Are our heuristics computationally efficient?}
Another downside of our method is the evaluation time.
For solved tasks, an average of only $3.42 \%$ of the evaluation time is spent on partition weight prediction thanks to our heavy GPU parallelization.
This places the evaluation speed bottleneck on feature extraction and abstraction heuristic computation.
In our implementation, feature generation and abstraction heuristic computation do not benefit from the available GPU.
This leads to a considerable amount of time being dedicated to those steps before acquiring the heuristic value.

\subsubsection{How informative is our learned heuristic compared to suboptimal partitions?}
Figure~\ref{fig:exp_nodes} compare the number of node expansions for the baselines and \hLCPRL{\FLG}{2}, our most effective learned configuration on this metric.  
The left-hand plot shows that our method requires fewer expansions than \hGZOCP for the majority of commonly solved tasks, which confirms that the learning is effective.
Plots with \hSCP and \hOCP show more mixed results. 
Nevertheless, a fair amount of tasks show a number of expanded nodes on par with \hOCP.

\section{Conclusion and Future Works}
We introduced a learning framework that infers admissible cost partitions for optimal planning.
By exploiting the equivalence between optimal cost partitioning and Lagrangian duals, our method replaces the expensive per‑state linear program with a learning model that predicts state‑aware partition weights.
The pipeline encodes a planning state and its pattern abstractions as a labelled graph, extracts action‑centric WL features and uses axial self‑attention with a softmax output to produce weights that satisfy the sum constraint by construction, guaranteeing admissibility.

Experiments on six IPC domains demonstrate that the learned partitions can reduce node expansions compared to the greedy zero‑one and saturated suboptimal baselines.
On the four domains where the model succeeds, coverage is competitive with the optimal cost partition.
However, the current implementation suffers from feature‑extraction and inference costs that make it slower than the suboptimal baselines and a lack of expressiveness on some domains, limiting its overall advantages.

Several directions can address the current limitations and extend our current work.
A more expressive graph representation could help our model generalize to larger problems.
Our architecture is currently trained on optimal cost partitions, but an optimal partition is only one of many possible ways to achieve an informative heuristic value.
The training pipeline could therefore be adjusted to incorporate the desired heuristic value directly as an additional label alongside the partition weights, so that the model learns to prefer partitions that not only mimic the optimal allocation but also maximize the resulting bound.
Another adjustment could be made to our way of exploring the state-space of our training instances to extract optimal partitions.
When the space and possible actions grows, the random walk might drift away from the optimal plan toward less relevant states and hinder training.
As we have noted, most of the evaluation time is not spent on model inference; speeding up feature extraction and abstraction heuristic computation could thus alleviate the main bottleneck. 
Both steps are currently implemented on CPU, whereas they are heavily parallelizable.
Moving them to the GPU could substantially reduce their runtime.
Finally, other strategies such as interval‑based online partitioning~\cite{seipp:21} could further improve speed if the predicted cost partition is sufficiently robust.

\section{Acknowledgements}
We thank the the anonymous reviewers for their helpful suggestions.
Sylvie Thi\'ebaux was funded by the Australian Research Council (ARC) under the Discovery Project grant DP220103815 and by the Artificial and Natural Intelligence Toulouse Institute (ANITI) under the grant agreement ANR-23-IACL-0002. Quentin Cappart received the support of the Natural Sciences and Engineering Research Council of Canada (RGPIN-2022-03964).

\bibliography{main}


\end{document}